\documentclass{article}
\usepackage{ijcai25}

\usepackage{times}
\usepackage{soul}
\usepackage{url}
\usepackage{booktabs}
\usepackage{multirow}
\usepackage[hidelinks]{hyperref}
\usepackage[small]{caption}
\usepackage{amsthm}
\usepackage{bm}
\usepackage{booktabs}
\usepackage[switch]{lineno}
\usepackage[T1]{fontenc}

\usepackage{subfig}
\usepackage{graphicx}
\usepackage[utf8]{inputenc}
\usepackage{array}
\usepackage{nccmath}
\usepackage{amsmath,bm}
\usepackage{amssymb}
\usepackage{mathrsfs} 
\usepackage{diagbox}
\usepackage{mathtools, nccmath, relsize}
\usepackage{enumitem}
\usepackage{makecell, multirow, tabularx}
\usepackage{algorithm}
\usepackage{algpseudocode}
\usepackage{graphics}

\newcolumntype{Y}{>{\centering\arraybackslash}X}
\usepackage{threeparttable}

\PassOptionsToPackage{table,xcdraw}{xcolor}

\usepackage{makecell}
\usepackage{colortbl}

\usepackage{calc}
\newcolumntype{C}[1]{>{\centering\arraybackslash}p{#1}}%
\newlength{\mycolwidth}
\newlength{\mycolwidthadd}
\newlength{\mytwocolwidthadd}
\newlength{\myfourcolwidth}
\newlength{\mythreecolwidth}
\newlength{\mytwocolwidth}
\usepackage{pifont}

\title{DeCo: Defect-Aware Modeling with Contrasting Matching for Optimizing Task Assignment in Online IC Testing}

\author{
Lo Pang-Yun Ting
\and
Yu-Hao Chiang
\and
Yi-Tung Tsai
\and
Hsu-Chao Lai
\And
Kun-Ta Chuang\\
\affiliations
Dept. of Computer Science and Information Engineering, National Cheng Kung University, Taiwan\\
\emails
\{lpyting, yhchiang, yttsai, hclai\}@netdb.csie.ncku.edu.tw,
\\
ktchuang@mail.ncku.edu.tw
}

\begin{document}
\maketitle

\begin{abstract}

In the semiconductor industry, integrated circuit (IC) processes play a vital role, as the rising complexity and market expectations necessitate improvements in yield. Identifying IC defects and assigning IC testing tasks to the right engineers improves efficiency and reduces losses. While current studies emphasize fault localization or defect classification, they overlook the integration of defect characteristics, historical failures, and the insights from engineer expertise, which restrains their effectiveness in improving IC handling.  To leverage AI for these challenges, we propose \emph{\textbf{DeCo}}, an innovative approach for optimizing task assignment in IC testing. \emph{DeCo} constructs a novel defect-aware graph from IC testing reports, capturing co-failure relationships to enhance defect differentiation, even with scarce defect data. Additionally, it formulates defect-aware representations for engineers and tasks, reinforced by local and global structure modeling on the defect-aware graph. Finally, a contrasting-based assignment mechanism pairs testing tasks with QA engineers by considering their skill level and current workload, thus promoting an equitable and efficient job dispatch. Experiments on a real-world dataset demonstrate that \emph{DeCo} achieves the highest task-handling success rates in different scenarios, exceeding 80\%, while also maintaining balanced workloads on both scarce or expanded defect data. Moreover, case studies reveal that \emph{DeCo} can assign tasks to potentially capable engineers, even for their unfamiliar defects, highlighting its potential as an AI-driven solution for the real-world IC failure analysis and task handling.

\end{abstract}
\section{Introduction}
\label{sec:intro}

\begin{table*}[h]
\small
\centering
\renewcommand{\arraystretch}{0.7}
\setlength{\tabcolsep}{3pt}
\begin{tabular}{llll}
\toprule[1.3pt]
File Category & Variable & Description & Example \\ \midrule
\multirow{5}{*}{RMA information} & $rma.id$ & 
The individual numeric identifier for the received RMA from customers. & 43219Q \\
 & $rma.start$ & The date (year, month, and day) on which an RMA is received. & 2020-11-25 \\
 & $rma.close$ & The date (year, month, and day) on which an RMA was completed. & 2020-12-20 \\
 & $rma.eng$ & The name of the engineer responsible for handling the RMA. & Teila \\
 & $defect.type$ & The primary defect type causing the IC failure in an RMA. & \#EIPD \\ \midrule
\multirow{4}{*}{ATE logs of an RMA} & $test.id$ & The individual identifier for the test item. & 1022 \\
 & $test.name$ & The name for the test item. & Ripple\_L04\_-VCC \\
 & $limit.min$ & The minimum value of the normal test range for this test item. & 0.000 dB \\
& $measure$ & The measured value of the test item. & 0.225 dB\\
 & $test.result$ & The test results of the test item. & passed \\

\bottomrule[1.3pt]
\end{tabular}
\caption{Overview of the data structure of an RMA and its ATE logs.}
\label{table:data_describe}
\end{table*}

As the demand for integrated circuits (ICs) continues to grow across numerous specialized sectors, it is essential in the semiconductor industry to minimize the time from production to shipment while maintaining quality at every stage. When manufacturers produce and deliver samples to customers, they receive numerous return material authorizations (RMAs)~\cite{10467052}. RMAs play a crucial role in the development of the next IC phase, as customer-reported issues are analyzed by the quality assurance (QA) and test verification teams. Addressing these issues directly affects IC design improvements and production efficiency, making RMA handling a key competitive factor.

\begin{figure}
\graphicspath{{figs/}}
\begin{center}
\includegraphics[width=0.49\textwidth]{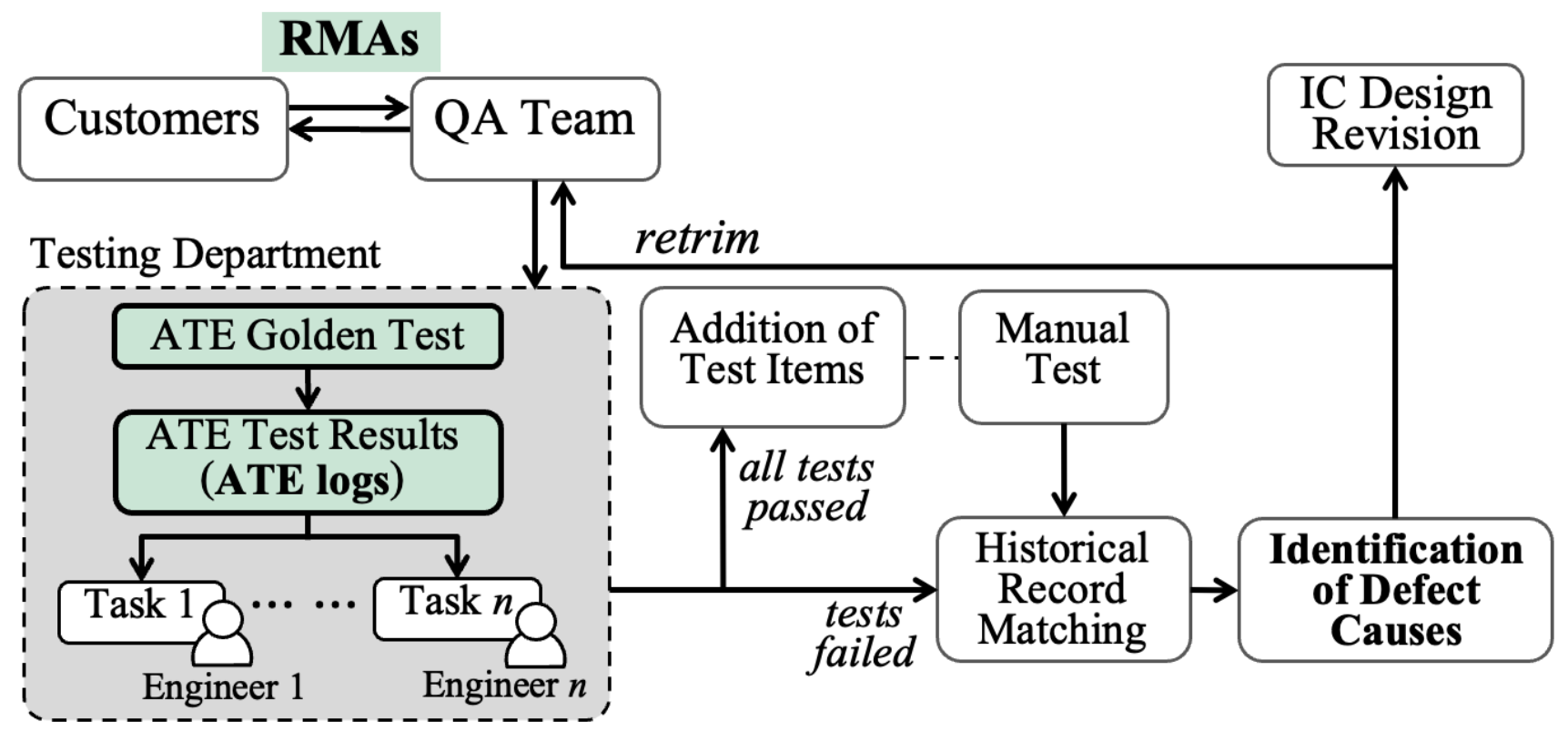}
\end{center}
\caption{The RMA processing flow.}
\label{action_struc}
\end{figure}

The RMA handling process is shown in Figure~\ref{action_struc}. In the testing department, RMA first undergoes golden version testing through the automatic test equipment (ATE)~\cite{1246545}, where each RMA is regarded as a unique task assigned to an engineer. The ATE generates a test report (ATE logs) containing results for various test items. Engineers analyze these reports to identify IC defect types, using historical RMA records and experience to diagnose the root cause. Once determined, they report findings to the QA team, aiding IC design improvements.

Current studies concentrate on automating fault localization within IC testing~\cite{9066890,10035984}, with the goal of pinpointing potential fault locations or defect types during manufacturing, or improving the detection yield during the design phase to reduce both cost and time~\cite{4369338,9790891}.
However, focusing solely on fault localization may lead to critical limitations. \ding{182} Specifically, certain defect types exhibit similar failure characteristics, and when defect data are scarce, identifying them becomes inherently challenging. As such, relying solely on automated fault localization is inadequate, as it does not utilize contextual insights from previous cases. Engineers with prior experience in similar failure scenarios are still often required to ensure accurate defect diagnosis. 
\ding{183} In additions, existing approaches overlook the structural relationships between defects, historical failure logs, and engineers' workload and expertise, which hinder effective RMA handling by leading to suboptimal task assignments and inefficient defect resolution. These limitations reveal a research gap: existing work focuses on defect detection, but lacks research on optimizing RMA handling to reduce manual diagnosis time and enhance task allocation efficiency as shown in Figure~\ref{action_struc}.

Addressing this gap is essential for reducing resolution time and enhancing overall IC testing workflows. To address the identified challenges and enhance RMA handling, we propose \textbf{DeCo} (\underline{\textbf{De}}fect-Aware Modeling with \underline{\textbf{Co}}ntrasting Matching), a novel framework designed to optimize RMA task assignment by maximizing task-handling success rates, while ensuring engineer workloads remain manageable.  Specifically, to overcome the first limitation, we design a \textbf{defect-aware graph} based on ATE logs of each RMA task, structuring co-failure relationships among RMAs. This graph structure encodes hidden dependencies among defects, enabling more accurate defect differentiation even when ATE logs for defect types are scarce. Additionally, we formulate \textbf{defect-aware representations} for engineers and RMA tasks based on their relationships with defect types. To further reinforce representations, we design \underline{local and global structure modeling} on the defect-aware graph, effectively capturing failure characteristics. These two modeling approaches can be deployed independently or integrated to provide a more comprehensive understanding of defect characteristics, directly addressing the second limitation. Finally, a \textbf{contrasting-based assignment} is introduced to match new RMA tasks with the most suitable engineers based on their representations and engineers' current workloads, ensuring optimal task assignment and improving overall efficiency in RMA handling.

Our key contributions are summarized as follows:

\begin{itemize}[leftmargin=*]
    \item We introduce \emph{DeCo}, a novel framework for optimizing RMA task assignment, maximizing task-handling success rates while balancing engineer workloads.
    \item To capture the failure characteristics of RMA tasks, we construct a defect-aware graph and design both local and global structure modeling to improve defect-aware task representations, using a contrastive-based mechanism to refine task-to-engineer assignments.

    \item We evaluate \emph{DeCo} on a real-world dataset under both scarce and expanded ATE log conditions. Results demonstrate that \emph{DeCo} achieves the highest success rate while keeping workloads manageable across different simulation scenarios, highlighting its potential as an AI-driven solution for intelligent 
    IC failure analysis and RMA handling.
\end{itemize}

\section{Preliminaries}

\subsection{Data Analysis}
\label{subsec:data_analysis}

\begin{figure}[t]
\graphicspath{{figs/}}
\begin{center}
\includegraphics[width=0.49\textwidth]{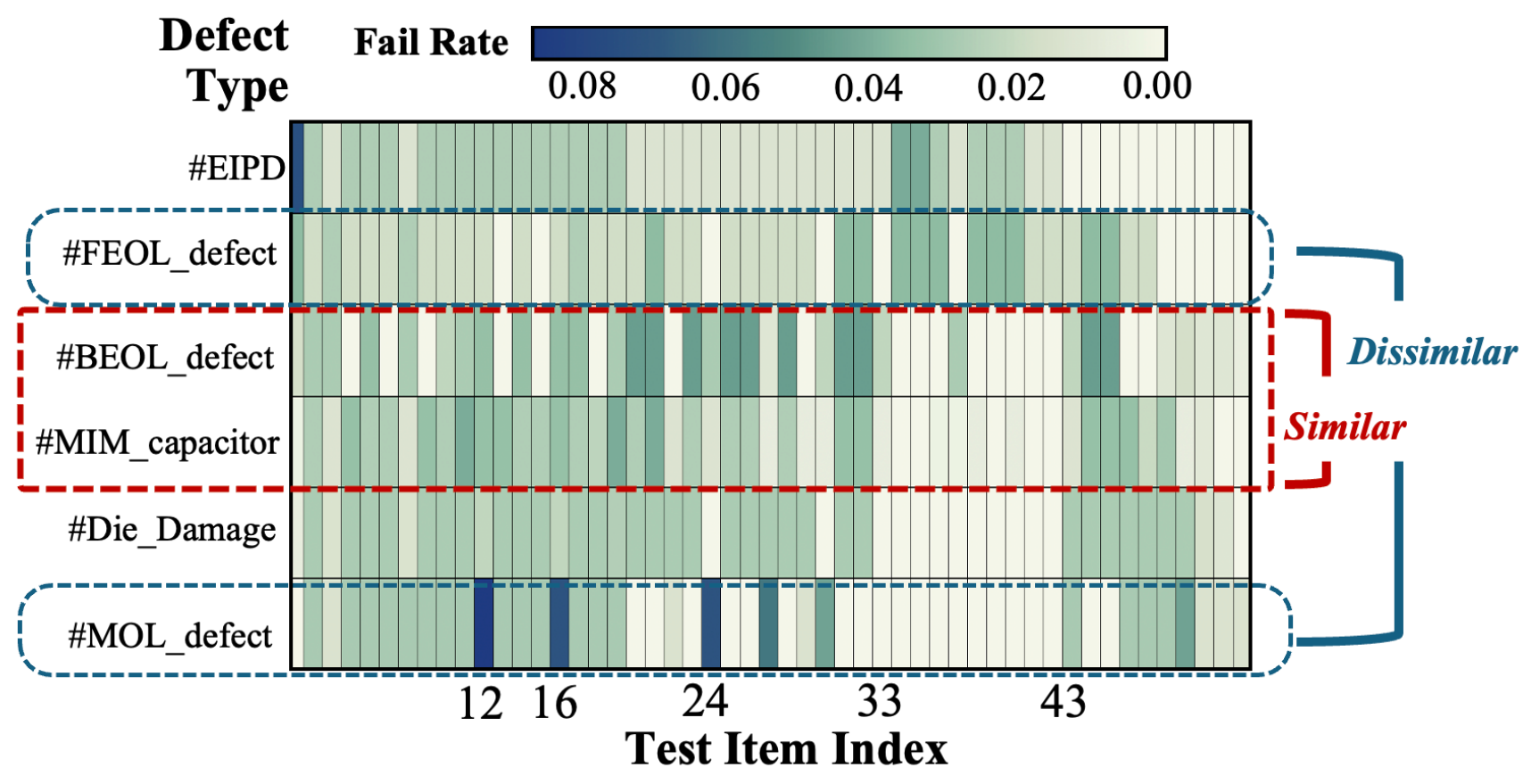}
\end{center}
\caption{Fail rates of each defect type of IC modules on different test items in ATE logs.}
\label{fig:data_observe}
\end{figure}

\subsubsection{Real-World Dataset}
\label{subsec:dataset}

For this study, we use real data from a semiconductor company, including RMAs for IC failures and corresponding ATE logs from 2020 to 2023, handled by four engineers. Table~\ref{table:data_describe} lists key data fields, with certain fields (e.g., $rma.id$, $rma.eng$, $test.id$) anonymized. Engineers categorize RMAs into $defect.type$ after analyzing ATE log results. Each ATE log contains multiple test items and their normal test range ($limit.min$). An RMA fails a test item if its measured value ($measure$) exceeds the normal range.

\subsubsection{Analysis on Failure Characteristics}
We perform a preliminary data analysis to validate that different defect types may share similar failure characteristics, as mentioned in Sec.~\ref{sec:intro}. Figure~\ref{fig:data_observe} shows the fail rate of each test name across defect types in all ATE logs. While fail rates vary, some defect types exhibit similar failure characteristics, whereas others differ significantly.

\noindent\underline{\textbf{Similar Characteristics.}} Defect types ``\#\textit{BEOL\_defect}'' and ``\#\textit{MIM\_capacitor}'' (highlighted by the red dotted line in Figure~\ref{fig:data_observe}) exhibit nearly zero fail rates between test indices 33 to 43 and share similar fail rate distributions in other regions, particularly in indices 24 to 33. ``\#\textit{MIM\_capacitor}'' refers to issues in metal-insulator-metal capacitors, such as capacitance deviations or dielectric defects, while ``\#\textit{BEOL\_defect}'' involves back-end-of-line manufacturing problems, including metal interconnect failures, via defects, or material delamination. The similarities between these two defect types arise from their shared origins in manufacturing challenges, such as material impurities, process misalignments, or contamination, which can lead to comparable failure characteristics.

\noindent\underline{\textbf{Dissimilar Characteristics.}}
Some defect types exhibit distinct failure characteristics. Specifically, ``\#\textit{FEOL\_defect}'' and ``\#\textit{MOL\_defect}'' (highlighted by the blue dotted line in Figure~\ref{fig:data_observe}) show substantial differences in fail rate distributions. ``\#\textit{FEOL\_defect}'' has higher fail rates at indices 30 to 43, while ``\#\textit{MOL\_defect}'' is significantly higher at indices 12, 16, and 24. This discrepancy arises because the type ``\#\textit{FEOL\_defect}'' occurs in the Front-End-Of-Line (FEOL) stage, affecting transistor performance, whereas ``\#\textit{MOL\_defect}'' originates in the Middle-Of-Line (MOL) stage, impacting signal transmission and interconnect resistance. As these defects arise at different manufacturing stages, their failure characteristics differ.

This finding shows that defect types vary in their associations with failure characteristics. Capturing these characteristics in ATE logs is crucial for assigning RMA tasks to suitable engineers. Accurate task matching not only enhances successful task handling but also reducing inefficiencies from misassignments, helping minimize the overall workload needed to complete all tasks.

\subsection{Problem Definition}

\noindent\textbf{Definition 1 (Task):} In this study, each RMA initiated by a customer is regarded as an unique task that requires handling. The set of tasks is denoted as $\mathcal{V}_{task}$.

\noindent\textbf{Problem Statement:} Given a set of tasks $\mathcal{V}_{task}$ and a set of engineers $E$, each task $t \in \mathcal{V}_{task}$ is associated with its ATE log testing results and defect type, and each engineer $e\in E$ has records of previously handled RMAs. Note that new RMA tasks arrive sequentially and are deemed successfully accomplished if they are finished \underline{prior to their designated close dates} (variable $rma.close$ in Table~\ref{table:data_describe}). Our goal is to assign new tasks to appropriate engineers to maximize the success rate while ensuring workloads remain manageable\footnote{``Success Rate'' and ``Workload'' are defined as the ratio of successfully accomplished tasks and the average number of tasks handled per engineer, respectively. For ease of presentation, details are described in Sec.~\ref{subsec:setup}.}.

\section{Proposed Method: DeCo}

The architecture of \emph{DeCo} comprises two main components. First, we formulate representations for engineers and RMA tasks based on their relationships with defect types. Simultaneously, local and global structure modeling are employed to better capture the failure characteristics in ATE logs (Sec.~\ref{sec:defect_represent}). Using these representations, the assignment of each engineer to a new RMA task is determined through a contrastive approach to identify the most appropriate task assignment plan (Sec.~\ref{sec:contrast_assign}).

\subsection{Defect-Aware Representation}
\label{sec:defect_represent}
\subsubsection{Engineer Representation}
To evaluate the ability of each engineer in handling different defect types of ICs, we first formulate the normalized representation $\bar{\textbf{e}}_i$  for each engineer $e_i\in E$, utilizing the data from RMAs they have previously handled. Given the defect type set $\mathcal{V}_{type}$, let $\mathcal{P}_{i,p}$ and $\mathcal{N}_{i,p}$ denote the average processing time and the count of defect type $p \in \mathcal{V}_{type}$ handled by engineer $e_i \in E$, respectively. An engineer $e_i$'s ability to handle defect type $p$ is defined as follows, where shorter processing time and more handling experience indicate higher ability:

\begin{equation}
\label{eq:ability}
{\small	
\text{ability}(i,p)= \text{avg}\biggl(1-\frac{\mathcal{P}_{i,p}}{\sum_{e_j}^{}\mathcal{P}_{j,p}/|E|}\:,\: \frac{\mathcal{N}_{i,p}}{\sum_{e_j}^{}\mathcal{N}_{j,p}/|E|}\biggr).
}
\end{equation}

Based on the defined ability $a_{i,p}$ of engineer $e_i$, the normalized defect-aware representation $\bar{\textbf{e}}_i \in \mathbb{R}^{|\mathcal{V}_{type}|}$ of engineer $e_i$ is formulated as follows:
\begin{equation}
\label{eq:engineer_repre}
{\small	
\bar{\mathbf{e}}_i = \left[ \frac{\text{ability}(i,p)}{\sum_{p'}^{}\text{ability}(i,p')} \right]_{p \in \mathcal{V}_{type}}.
}
\end{equation}

\subsubsection{Task Representation}

To preserve ATE log information and represent new tasks, we construct a defect-aware graph based on ATE logs and \textit{encode each node within it} to capture failure characteristics.

\begin{figure}
\graphicspath{{figs/}}
\begin{center}
\includegraphics[width=0.485\textwidth]{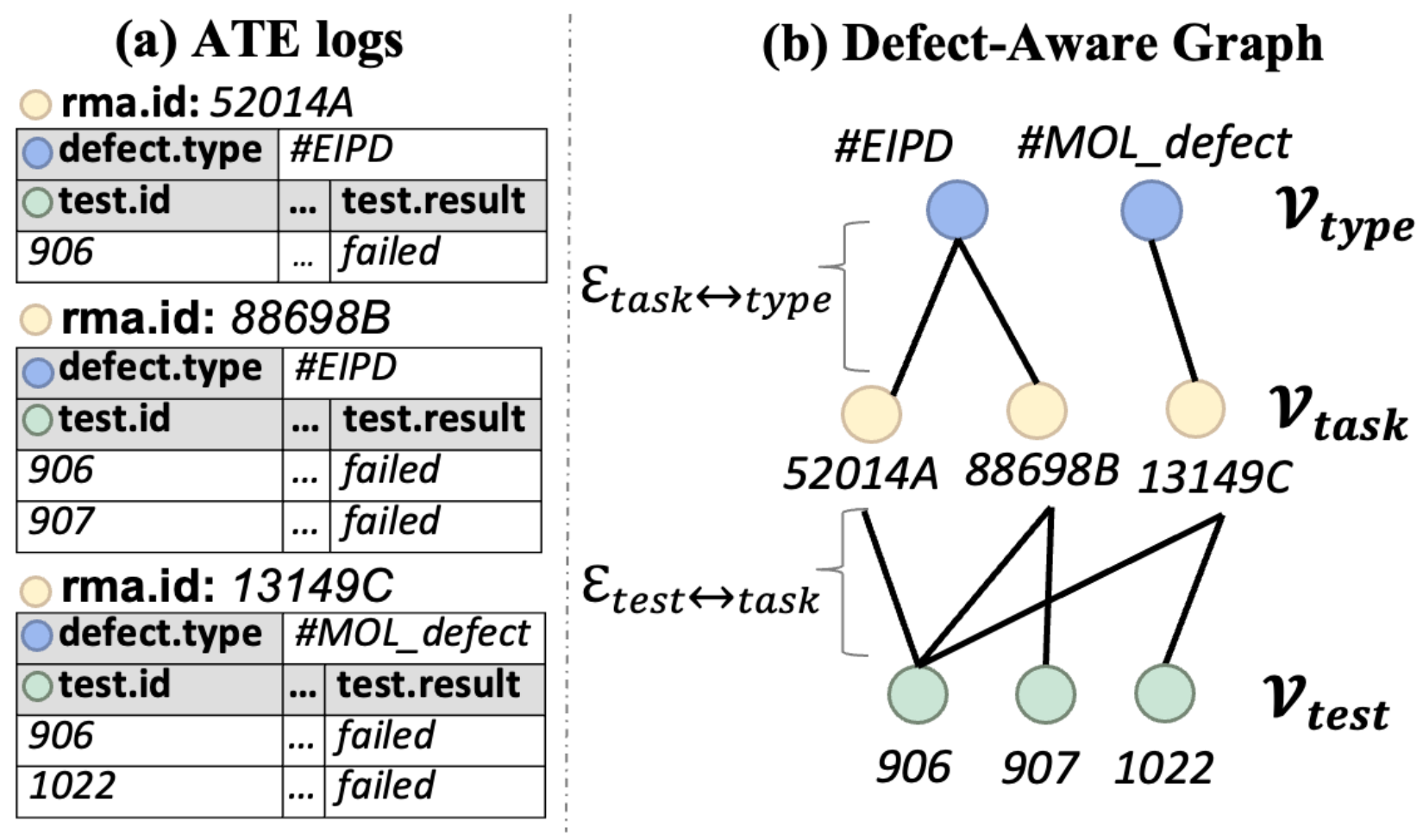}
\end{center}
\caption{Defect-aware graph construction from ATE logs.}
\label{fig:defect_graph}
\end{figure}

\noindent\textbf{Definition 2 (Defect-Aware Graph):} 
Given the task set $\mathcal{V}_{task}$, defect type set $\mathcal{V}_{type}$, and the test item set $\mathcal{V}_{test}$, the relationships between these entities are defined as follows. If a test item $m \in \mathcal{V}_{test}$ fails in the ATE logs of task $t \in \mathcal{V}_{task}$, an edge $\varepsilon_{m\leftrightarrow t}$ is formed, defining the edge set $\mathcal{E}_{test \leftrightarrow task}=\{\varepsilon_{m\leftrightarrow t} \mid m \in \mathcal{V}_{test}, t \in \mathcal{V}_{task}\}$. If a task $t \in \mathcal{V}_{task}$ belongs to defect type $p \in \mathcal{V}_{type}$, an edge $\varepsilon_{t\leftrightarrow p}$ is formed, defining the edge set $\mathcal{E}_{task \leftrightarrow type}=\{\varepsilon_{t\leftrightarrow p} \mid t \in \mathcal{V}_{task}, p \in \mathcal{V}_{type}\}$. Therefore, the defect-aware graph $\mathcal{G}$ is formulated as follows:

\begin{equation}
\label{eq:graph}
{\small	
\mathcal{G}=(\mathcal{V},\mathcal{E},\mathcal{X})\begin{cases}
\mathcal{V}=\mathcal{V}_{test}\cup \mathcal{V}_{task} \cup \mathcal{V}_{type}\:\: (nodes)
\\ 
\mathcal{E}=\mathcal{E}_{test \leftrightarrow task}\cup \mathcal{E}_{task \leftrightarrow type}\:\: (edges)
\\ 
\mathcal{X}=\{x_v | v\in \mathcal{V}\}\:\: (attributes)
\end{cases},
}
\end{equation}
where $\mathcal{X}$ is an attribute set, and each $x_v \in \mathcal{X}$ represents a node attribute of ``\textit{test item}'', ``\textit{task}'' or ``\textit{defect type}''. Figure~\ref{fig:defect_graph} illustrates an example of graph construction.

The observation in Figure~\ref{fig:data_observe} highlights that different defect types exhibit varying degrees of correlation in their failure characteristics. To effectively capture these characteristics and encode each node in the defect-aware graph $\mathcal{G}$, we design two types of modeling approaches: \textit{local structure modeling} and \textit{global structure modeling}. These approaches \underline{can be applied individually or in combination}. Generally, integrating both improves performance by capturing diverse structural information. In some cases, such as when ATE logs are extensive, global structure modeling alone better captures high-level defect relationships since it leverages broader connections. Conversely, local structure modeling alone may yield better results. By flexibly applying local and global structural information, \emph{DeCo} can provide a more comprehensive representation of defect-aware relationships.

\noindent\textbf{\textit{Local structure modeling.}}
For local structure modeling, we represent the characteristics of each node in the defect-aware graph $\mathcal{G}$ by exploring its local neighborhood. This allows us to encode fine-grained, proximity-based relationships among test items, tasks, and defect types. To achieve this, we redesign the classic model DeepWalk~\cite{perozzi2014deepwalk}, which learns node embeddings by generating random walks over the graph and optimizing node co-occurrence within local neighborhoods. Also, we extend the skip-gram architecture~\cite{node2vec,w2v} by considering the \underline{second-order proximity}~\cite{Tang2015LINELI} of each node in defect-aware graph $\mathcal{G}=(\mathcal{V}, \mathcal{E}, \mathcal{X})$ to better capture structural proximity. The objective function of local structure modeling is designed as follows:

\begin{equation}
\label{objective_local}
\mathcal{O}_{local}=-\log Z_{u} + \sum_{v\in N(u)}^{} \Bigl(1+\omega_{v,u}\Bigr) \Bigl(\psi(v)\cdot \psi(u) \Bigr),
\end{equation}

\begin{equation}
\label{weight_sim}
{\small	
\omega_{v,u}\hspace{-0.1cm} =\hspace{-0.1cm}\begin{cases}
 J\bigl(N(v),N(u)\bigr)& \text{\hspace{-0.2cm}, if } v\neq u \text{ and } v, u \in \mathcal{V}_{task} \\ 
 0& \text{\hspace{-0.2cm}, otherwise }
\end{cases},
}
\end{equation}
where $Z_u = \sum_{u\in \mathcal{V}} \exp(\psi(v) \cdot \psi(u))$, and $N(u)$ refers to th set of nodes in the random walk starting from $u$. The function $J(N(v), N(u))$ denotes the Jaccard similarity between $N(v)\cap \mathcal{V}_{test}$ and $N(u)\cap \mathcal{V}_{test}$. $\psi(v) \in \mathbb{R}^d$ and $\psi(u) \in \mathbb{R}^d$ are embedding vectors of nodes $v$ and $u$, respectively.

\noindent\textbf{\textit{Global structure modeling.}} For global structure modeling, unlike direct neighborhood relationships, we construct metapaths to traverse multiple intermediate nodes, uncovering \underline{high-level relational dependencies}. A metapath is an ordered sequence of node types and edge types defined on the network schema. These structures are specifically designed to reveal high-level relational dependencies. The metapaths used in our work are defined as:

\noindent\textbf{Definition 3 (Metapath):} We define a metapath $P$ as a path in the form of $u(n_0)\overset{e_1}{\rightarrow}n_1\overset{e_2}{\rightarrow}...\overset{e_l}{\rightarrow}v(n_{l})$, where $e_j \in \mathcal{E}$, and $u$ and $v$ belong to the same node set, i.e., both are from test items ($\mathcal{V}_{test}$), tasks ($\mathcal{V}_{task}$), or defect types ($\mathcal{V}_{type}$). The intermediate nodes $\{n_1, ..., n_{l-1}\}$ belong to the remaining node sets. For example, if $u, v \in \mathcal{V}_{type}$, the intermediate nodes satisfy $\{n_1, \dots, n_{l-1}\} \subset \mathcal{V}_{task} \cup \mathcal{V}_{test}$. The metapath describes a composite relationship $\pi = e_1 \circ n_1 \circ e_2 \circ  ... \circ e_l$ between node $u$ and $v$, where $\circ$ denote the composition operator that sequentially connects edges and intermediate nodes in $\mathcal{G}$. Therefore, a metapath structure can be represented in a triplet form $P = (u, \pi, v)$.

To strengthen these structured dependencies, we create a triplet-based representation and employ a margin-based objective to grasp the interconnections between the source node $u$ and the target node $v$ within a metapath $P=(u, \pi, v)$. This approach facilitates the learning of representations for various nodes in the defect-aware graph.
Let $\mathcal{P}^{\le l}_{u,v}$ denote all metapaths from $u$ to $v$ with a length of at most $l \in \mathbb{N}$. The objective function of global structure modeling is defined as:

\begin{equation}
\label{objective_global}
\mathcal{O}_{global}=- \frac{1}{|\mathcal{P}^{\le l}_{u,v}|} \sum_{P=(u,\pi, v) \in \mathcal{P}^{\le l}_{u,v}}^{} L_P,
\end{equation}

\begin{equation}
\label{eq:kge_loss}
{\small	
L_P = \sum_{P' \in \{(u', \pi, v') \mid u' \neq u \text{ or } v' \neq v\}} \Bigl[E_P + \xi - E_{P'}\Bigr]_+,
}
\end{equation}
where $P'$ is the set of negative samples generated based on metapath $P$, and $E_P$ is a scoring function that evaluates the plausibility of a metapath $P$. This function is derived from the concept of knowledge graph embedding, ensuring that the metapath (triplet structure) captures meaningful structural relationships. We employ the HolE model~\cite{nickel2016holographic} as the scoring function for its effectiveness in capturing complex relational triplet structures. Here, $[\mu]_{+}$ denotes $max(0,\mu)$, and $\xi > 0$ is a hyperparameter of margin.

The objective function for integrating both modeling approaches can be easily derived as:

\begin{equation}
\label{skip-gram-kge}
{\small	
\begin{aligned}
&\hspace*{0cm}
\max_{\psi } (\mathcal{O}_{local}) + \lambda(O_{global})
\\
&\hspace*{0cm}
= \max_{\psi }\sum_{u\in \mathcal{V}}^{}\bigg\{\underbrace{-\log Z_{u} + \sum_{v\in N(u)}^{} \bigg[\Bigl(1+\omega_{v,u}\Bigr) \Bigl(\psi(v)\cdot \psi(u) \Bigr)}_{\text{local structure}}
\\
&\hspace*{2cm}
-  \lambda\cdot\underbrace{\frac{1}{|\mathcal{P}^{\le l}_{u,v}|} \sum_{P \in \mathcal{P}^{\le l}_{u,v}}^{} \sum_{P'} \Bigl[E_P + \xi - E_{P'}\Bigr]_+}_{\text{global structure}}\bigg]\bigg\},
\end{aligned}
}
\end{equation}
where $\lambda  > 0$ is a hyperparameter balancing both modeling. Finally, the representation of each test item, task and defect type in the defect-aware graph can be obtained.

Based on Eq.~\ref{skip-gram-kge}, we can formulate the representation of a new task $t_k$ ($\notin \mathcal{V}_{task}$) by considering the failed test items in its ATE logs. Let $\mathcal{F}_{k} \subset \mathcal{V}_{test}$ includes the failed test items in the ATE logs of new task $t_k$. The probability of task $t_k$ belongs to defect type $p \in \mathcal{V}_{type}$ is estimated as follows:

\begin{equation}
\label{eq:sim}
{\small	
\text{prob}(k,p) = sim\biggl(\frac{1}{|\mathcal{F}_{k}|} \sum_{f\in \mathcal{F}_{k}}^{} \psi(f), \: \psi(p) \biggr),
}
\end{equation}
where $\psi(\cdot)$ represents the embedding vector of a node. The function $sim(\cdot, \cdot)$ represents the cosine similarity. Subsequently, the normalized defect-aware representation $\bar{\mathbf{t}}_k \in \mathbb{R}^{|\mathcal{V}_{type}|}$ of new task $t_k$ is designed as follows:

\begin{equation}
\label{eq:task_repre}
{\small	
\bar{\mathbf{t}}_k = \left[ \frac{\text{prob}(k,p)}{\sum_{p'}^{}\text{prob}(k,p')} \right]_{p \in \mathcal{V}_{type}}.
}
\end{equation}

\subsection{Contrasting-Based Assignment}
\label{sec:contrast_assign}

After obtaining the defect-aware representation $\bar{\mathbf{e}}_i$ (Eq.~\ref{eq:engineer_repre}) and $\bar{\mathbf{t}}_k$ (Eq.~\ref{eq:task_repre}) for each engineer $e_i$ and newly received task $t_k$, we aim to assign the most appropriate engineer to handle the new task $t_k$. Here, we design the \textbf{assignment score} $\mu_{i,k}$ to quantify the suitability of each engineer $e_i\in E$ for new task $t_k$. Inspired by contrastive decoding approaches~\cite{li2022contrastive,chuang2023dola}, the assignment score $\mu_{i,k}$ is computed by contrasting information across representations $\bar{\mathbf{t}}_k$ and $\bar{\mathbf{e}}_i$ to refine the task-to-engineer assignment process, as designed as follows:

\begin{equation}
\label{eq:orm_match_score}
{\small	
\mu_{i,k}=\sum_{p\in \mathcal{V}_{type}}^{} \mathcal{S}(e_i, t_k,p),
}
\end{equation}

\begin{equation}
\label{eq:unnorm_match_score}
{\small	
\mathcal{S}(e_i, t_k,p)=\begin{cases}
 \log\cfrac{\bar{\textbf{e}}_{i}(p)+\epsilon}{\bar{\textbf{t}}_{k}(p)+\epsilon}& \text{, if } N_{i,p}\neq 0 \text{ and } \frac{w_i^{curr}}{w_i^{avg}} < 1 \\ 
 -\infty& \text{, otherwise }
\end{cases},
}
\end{equation}
where $\epsilon>0$ is small constant introduced to prevent division by zero. $\bar{\textbf{e}}_i(p) = e_{i,p}$ represents the ability of engineer $e_i$ for defect type $p$ (Eq.~\ref{eq:ability}, Eq.~\ref{eq:engineer_repre}). $\bar{\textbf{t}}_k(p) = s_{k,p}$ indicates the probability that new task $t_k$ belongs to defect type $p$ (Eq.~\ref{eq:sim}, Eq.~\ref{eq:task_repre}). $N_{i,p}$ denotes the number of tasks of defect type $p$ handled by engineer $e_i$. 

Specifically, to prevent excessive workload on engineers, we introduce a workload constraint. The values $w_i^{curr}$ and $w_i^{avg}$ represent the current number of tasks handled by engineer $e_i$ and $e_i$'s average historical workload, respectively. The condition $w_i^{curr}/w_i^{avg} < 1$ ensures that engineers with lighter workloads are prioritized.

Finally, according to the assignment score $\mu_{i,k}$ for each engineer $e_i\in E$ and the new task $t_k$, we assign task $t_k$ to the engineer $\hat{e}$ with the \underline{highest score}, ensuring that the selected engineer has sufficient capacity and a manageable workload to handle the new task effectively.

\section{Experiments}

\subsection{Experimental Setup}
\label{subsec:setup}
\noindent\textbf{Dataset and Preprocessing.} We use a real-world ATE dataset collected from a semiconductor company, as described in Sec.~\ref{subsec:dataset}. To evaluate methods under both scarce and informative defect data conditions, we consider the original dataset as scarce and create an expanded dataset by increasing the number of engineers, defect types, RMA logs and other relevant factors. The statistics of the scarce (original) dataset and the expanded dataset are presented in Table~\ref{table:dataset}. The first 75\% of the total RMA data is used for training, while the remaining 25\% is reserved for testing.

\begin{table}[b]
\small
\centering
\renewcommand{\arraystretch}{0.68}
\setlength{\tabcolsep}{2.2pt}
\begin{tabular}{c|c|c|c|c|c}
\toprule[1pt]
Datasets & \# Testers &  \thead{\#Test \\ items} & \thead{\#Defect \\ types}  & \thead{\# ATE logs} & \thead{Avg. handling \\ days} \\ \midrule
\textbf{ATE}$_{\text{scar}}$ & 4 & 1,140 & 8 & 234 & 38.3\\
\textbf{ATE}$_{\text{exp}}$ & 26 & 1,703 & 14 & 3,190 &  41.1\\ 
\bottomrule[1pt]
\end{tabular}
\caption{Dataset statistics.}
\label{table:dataset}
\end{table}

\noindent\textbf{Baselines.} 
We compare \emph{DeCo} with the following baselines: $\textbf{LowestWL}$, $\textbf{MostEXP}$, $\textbf{CF}$, as well as our variant $\textbf{CA}$. LowestWL and MostEXP are widely used strategies in industries.

\begin{itemize}[leftmargin=*]

\item \textbf{LowestWL} assigns each new task to the engineer with the lowest current workload.

\item \textbf{MostEXP} assigns each new task to the engineer who has handled the most tasks with similar failed test items, considering them the most experienced.

\item \textbf{CF} employs collaborative filtering (CF). An engineer-task matrix from past RMA records calculates suitability scores. Similarity between a new task and past tasks uses Word2Vec embeddings of failed test items, averaging these embeddings to represent each task. An engineer's suitability for a new task is a weighted sum of past task scores, with weights from task cosine similarity.

\item \textbf{CA} is a variant of \emph{DeCo} retaining only \underline{c}ontrast-based \underline{a}ssignment (remove defect-aware graphs). The new task's representation uses cosine similarity between fail rates of test items in ATE logs and those of each defect type.
\end{itemize}

We also evaluate our method with local, global, and combined structure modeling, denoted as $\textbf{\emph{DeCo}}_{\text{local}}$, $\textbf{\emph{DeCo}}_{\text{global}}$, and $\textbf{\emph{DeCo}}$, respectively.

\begin{table*}[h]
\footnotesize	
\centering
\renewcommand{\arraystretch}{0.65}
\setlength{\tabcolsep}{5pt}
\begin{tabular}{l|c|ccccccc}
\toprule[1.3pt]
\multicolumn{9}{c}{Dataset $\textbf{ATE}_{\text{scar}}$} \\ 
\midrule[1.3pt]
\multirow{2}{*}{\textbf{Metric}}  & \multirow{2}{*}{\textbf{$\kappa$}} &  \multirow{2}{*}{LowestWL} & \multirow{2}{*}{MostEXP} & \multirow{2}{*}{CF} & \multirow{2}{*}{CA}  &  \multicolumn{3}{c}{\textit{\textbf{Ours}}} \\ \cmidrule(rl){7-9} 
&  &  &  &  &  &  \textbf{\emph{DeCo}}$_{\text{local}}$ & \textbf{\emph{DeCo}}$_{\text{global}}$ & \textbf{\emph{DeCo}} \\ \midrule \midrule
 Success(\%) ($\uparrow$) &  \multirow{2}{*}{1}  & 69.33$\pm$7.60 & 48.66$\pm$11.69 & 64.00$\pm$2.79 & 76.67$\pm$1.82 & 82.66$\pm$4.34 & 82.00$\pm$4.47 & $\bm{86.00\pm2.79}$ \\
Workload ($\downarrow$) && \cellcolor[gray]{0.82}{1.38$\pm$0.10} & 2.21$\pm$0.51 & \cellcolor[gray]{0.82}{1.66$\pm$0.11} & \cellcolor[gray]{0.82}{1.60$\pm$0.05} & \cellcolor[gray]{0.82}{1.66$\pm$0.06} & 1.71$\pm$0.08 & \cellcolor[gray]{0.82}{1.69$\pm$0.08} \\
\midrule[0.1pt]
 Success(\%) ($\uparrow$) &  \multirow{2}{*}{2}  & 68.00$\pm$5.58 & 48.66$\pm$14.45 & 64.66$\pm$2.98 & 74.49$\pm$2.61 & $\bm{84.66\pm1.82}$ & 80.00$\pm$4.71 & 83.99$\pm$3.65\\
Workload ($\downarrow$)  && \cellcolor[gray]{0.82}{1.45$\pm$0.17} & 2.19$\pm$0.53 & \cellcolor[gray]{0.82}{1.63$\pm$0.03} & \cellcolor[gray]{0.82}{1.54$\pm$0.15} & \cellcolor[gray]{0.82}{1.62$\pm$0.20} & 1.79$\pm$0.12 & \cellcolor[gray]{0.82}{1.68$\pm$0.10} \\
\midrule[0.1pt]
 Success(\%) ($\uparrow$) &  \multirow{2}{*}{5}  & 74.66$\pm$5.05 & 48.66$\pm$14.45 & 61.33$\pm$2.97 & 73.33$\pm$1.01 & 88.00$\pm$1.62 & 82.66$\pm$4.34 & $\bm{88.01\pm1.82}$ \\
Workload ($\downarrow$)  && \cellcolor[gray]{0.82}{1.36$\pm$0.18} & 2.18$\pm$0.53 & 1.63$\pm$0.07 & \cellcolor[gray]{0.82}{1.42$\pm$0.07} & \cellcolor[gray]{0.82}{1.50$\pm$0.14} & 1.62$\pm$ 0.16 &\cellcolor[gray]{0.82}{1.50$\pm$0.14} \\
\midrule[1.3pt]
\multicolumn{9}{c}{Dataset $\textbf{ATE}_{\text{exp}}$} \\ 
\midrule[1.3pt]
\multirow{2}{*}{\textbf{Metric}}  & \multirow{2}{*}{\textbf{$\kappa$}} &  \multirow{2}{*}{LowestWL} & \multirow{2}{*}{MostEXP} & \multirow{2}{*}{CF} & \multirow{2}{*}{CA}  &  \multicolumn{3}{c}{\textit{\textbf{Ours}}} \\ \cmidrule(rl){7-9} 
&  &  &  &  &  &   \textbf{\emph{DeCo}}$_{\text{local}}$ & \textbf{\emph{DeCo}}$_{\text{global}}$ & \textbf{\emph{DeCo}} \\ \midrule \midrule
 Success(\%) ($\uparrow$) &  \multirow{2}{*}{1}  & 76.09$\pm$1.94 & 13.74$\pm$11.06 & 48.35$\pm$2.12 & 78.37$\pm$2.24 & 78.63$\pm$2.28 & 78.06$\pm$1.69 & $\bm{79.32\pm1.19}$ \\
Workload ($\downarrow$)  && \cellcolor[gray]{0.82}{0.65$\pm$0.01} & 1.42$\pm$0.34 & \cellcolor[gray]{0.82}{0.70$\pm$0.01} & \cellcolor[gray]{0.82}{0.55$\pm$0.01} & \cellcolor[gray]{0.82}{0.70$\pm$0.06} & \cellcolor[gray]{0.82}{0.74$\pm$0.06} & \cellcolor[gray]{0.82}{0.68$\pm$0.04} \\
\midrule[0.1pt]
 Success(\%) ($\uparrow$) &  \multirow{2}{*}{2}  & 84.81$\pm$1.36 & 13.24$\pm$11.10 & 60.42$\pm$6.74 & 88.27$\pm$2.12 & $\bm{90.08\pm1.17}$ & 89.18$\pm$1.77 & 89.52$\pm$2.13\\
Workload ($\downarrow$)  && \cellcolor[gray]{0.82}{0.58$\pm$0.01} & 1.23$\pm$0.63 & 0.67$\pm$0.04 & \cellcolor[gray]{0.82}{0.48$\pm$0.01} & \cellcolor[gray]{0.82}{0.50$\pm$0.04} & \cellcolor[gray]{0.82}{0.49$\pm$0.04} & \cellcolor[gray]{0.82}{0.52$\pm$0.07}  \\
\midrule[0.1pt]
 Success(\%) ($\uparrow$) &  \multirow{2}{*}{5} & 89.38$\pm$0.58 & 13.21$\pm$11.25 & 70.36$\pm$3.59 & 95.07$\pm$1.12 & 95.13$\pm$1.31 & $\bm{95.83\pm0.71}$ & 95.34$\pm$0.83\\
Workload ($\downarrow$)  && \cellcolor[gray]{0.82}{0.33$\pm$0.01} & 1.44$\pm$0.03 & 0.62$\pm$0.03 & \cellcolor[gray]{0.82}{0.40$\pm$0.34} & \cellcolor[gray]{0.82}{0.41$\pm$0.02} & \cellcolor[gray]{0.82}{0.39$\pm$0.01} & \cellcolor[gray]{0.82}{0.41$\pm$0.01}\\
\bottomrule[1pt]
\end{tabular}
\caption{Performance comparison on datasets $\textbf{ATE}_{\text{scar}}$ and $\textbf{ATE}_{\text{exp}}$. The results are reported as the mean $\pm$ standard deviation over 10 runs. The best success rate (abbreviated as ``Success'') results are highlighted in \textbf{bold}. Cells in gray indicate workload ratios (abbreviated as ``Workload'') that are better than the average performance across methods.}
\label{table:assign_comparison}
\end{table*}

\noindent\textbf{Evaluation Metrics.} Our evaluation relies on two metrics:
\begin{itemize}[leftmargin=*]
\item \textbf{Success}: The \textit{success rate} (\%) is the ratio of tasks successfully completed by engineers, estimated as $|\mathcal{N}_{success}|/|\mathcal{N}_{total}|$, where $\mathcal{N}_{success}$ is the set of tasks completed before their close dates, and $\mathcal{N}_{total}$ includes all new tasks that need to be handled.

\item \textbf{Workload}: The metric reports the average \textit{daily workload} per engineer, with lower values being better. The daily workload of each engineer is estimated as $w_i^t/w_i^{avg}$, where $w_i^t$ is the number of tasks handled by engineer $e_i$ on day $t$, and $w_i^{avg}$ is their average daily task count. 
\end{itemize}

\noindent\textbf{Engineer Processing Time Simulation.} To simulate the time required for $e_i$ to handle the new task $t_k$, we define an influence factor $x$ based on the current workload and experience. It is estimated as: $x=(w_i^{curr}/w_i^{avg}) + (N_{i,p}/N^{avg}_i)$, where $w_i^{curr}$ and $w_i^{avg}$ are the current and average number of tasks handled by $e_i$, respectively. Similarly, $N_{i,k}$ is the number of times $e_i$ has handled tasks of the same defect type as $t_k$, and $N^{avg}_i$ is the average number of times an engineer handles each defect type. Given the influence factor $x$, the time $\mathcal{T}_{i,k}$ required for $e_i$ to handle the new task $t_k$ is as:

\begin{equation}
\label{eq:process_simulate}
\mathcal{T}_{i,k}=\mathcal{T}^{\text{avg}}_{i,k}+\sigma (\kappa\cdot x - 0.5)\Delta \mathcal{T}_i,
\end{equation}
where $\mathcal{T}^{avg}_{i,k}$ is the average time $e_i$ takes to handle tasks of the same defect type as $t_k$, and $\Delta \mathcal{T}_i$ is the difference between the longest and shortest handling times for $e_i$. \underline{The parameter $\kappa > 0$ controls the impact of the influence} \underline{factor $x$}, with a larger $\kappa$ means a greater influence of workload and experience on the required time for handling $t_k$.

\noindent\textbf{Implementation details.} For local structure modeling, the walk length and number of walks are set to 20 and 10, respectively. For global structure modeling, the metapath length $l$ is set to 5 (Eq.~\ref{objective_global}), and the margin $\xi$ is set to 1 (Eq.~\ref{eq:kge_loss}). In Eq.~\ref{skip-gram-kge}, $\lambda$ is set to 1, the embedding dimension to 128, and the training epoch to 50. All experiments are conducted on a system with 12 CPU cores and 64GB RAM, running CUDA 12.1.

\subsection{Results and Analysis}

\noindent\textbf{Comparison Results.} 
Table~\ref{table:assign_comparison} presents the comparative results under different engineer processing time settings, with $\kappa={1, 2, 5}$ (Eq.~\ref{eq:process_simulate}). Key observations include:

\noindent\underline{\textit{Effectiveness.}} Our framework outperforms all baselines in success rate, with the second-highest success rate also achieved by \emph{DeCo} variants (\emph{\textbf{DeCo}}${_\text{local}}$, \emph{\textbf{DeCo}}${_\text{global}}$, or \emph{\textbf{DeCo}}). Meanwhile, it maintains a competitive workload, demonstrating the effectiveness of capturing failure characteristics from the defect-aware graph. Although LowestWL and CA perform similarly to our method on the \textbf{ATE}$_{\text{exp}}$ dataset, their success rates remain lower across different $\kappa$ values, while their workload is only slightly lower.

\noindent\underline{\textit{Generalization.}} 
\emph{DeCo} achieves the highest success rate while keeping workloads manageable on both datasets. In \textbf{ATE}$_{\text{scar}}$, it significantly outperforms baselines, showing its applicability with limited data. In \textbf{ATE}$_{\text{exp}}$, although \emph{DeCo}$_{\text{local}}$, \emph{DeCo}$_{\text{global}}$, and \emph{DeCo} perform similarly, an interesting trend emerges: when more ATE logs are available (\textbf{ATE}$_{\text{exp}}$) and the engineer processing time is more affected by workload and experience ($\kappa = 5$), global structure modeling alone achieves a higher success rate. This is because it captures broader task relationships and engineer expertise, leading better task assignments when data is abundant and processing time depends more on workload and experience.

\begin{figure}[t]
\graphicspath{{figs/}}
\begin{center}
\includegraphics[width=0.43\textwidth]{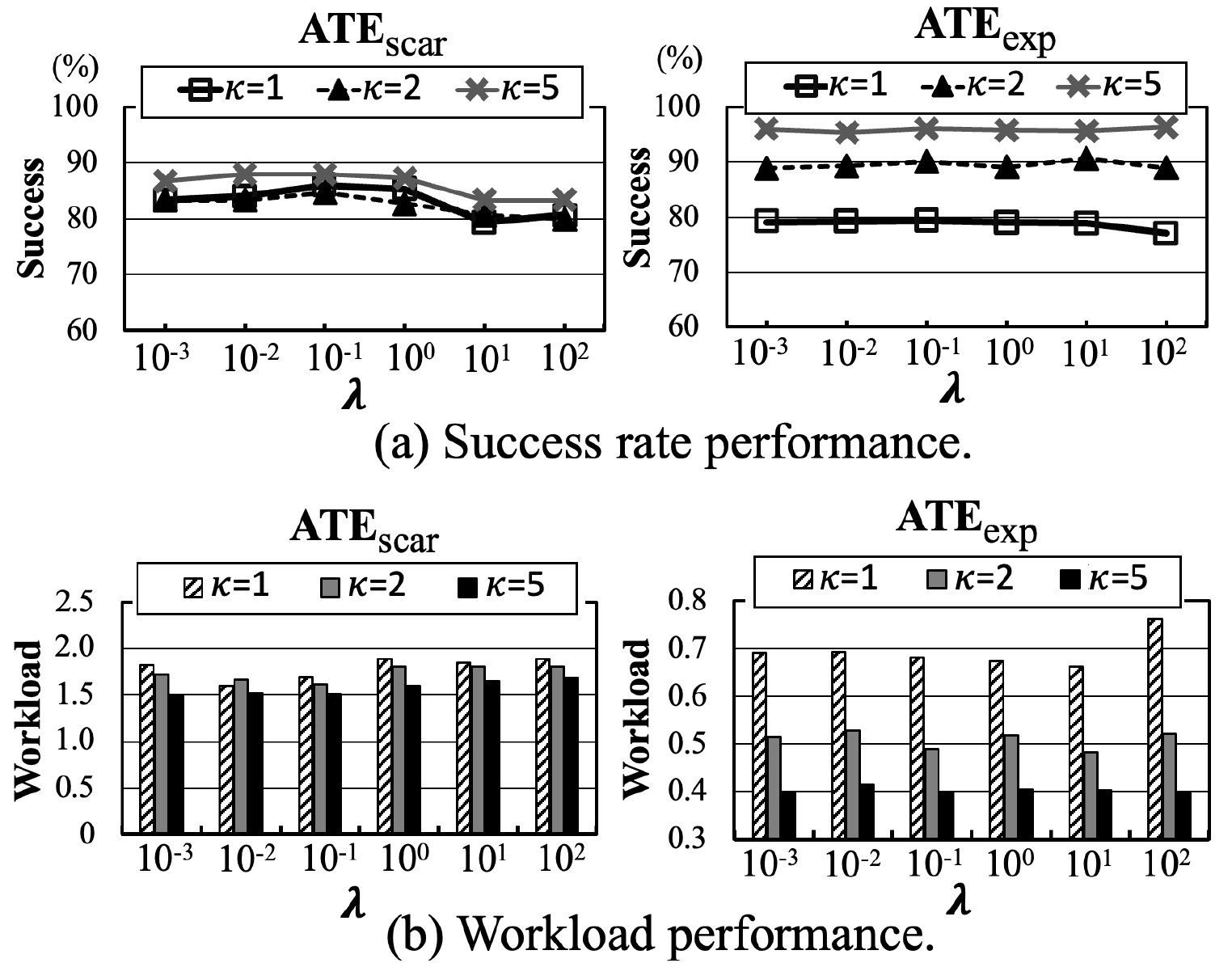}
\end{center}
\caption{The performance of \emph{DeCo} with different values of $\lambda$.}
\label{fig:global_effect}
\end{figure}

\noindent\textbf{Performance Analysis of Local and Global Modeling Integration.} 
To evaluate the impact of integrating local and global structure modeling in \emph{DeCo}, we vary the $\lambda$ value (Eq.~\ref{skip-gram-kge}) to control the degree of integration, as shown in Figure~\ref{fig:global_effect}. In dataset \textbf{ATE}$_{\text{scar}}$, setting $\lambda$ around $10^{-2}$ and $10^{-1}$ (where global modeling has less impact than local modeling) achieves the highest success rate and lowest workload. In dataset \textbf{ATE}$_{\text{exp}}$, the success rate remains stable across different $\lambda$ values, but setting $\lambda$ to $10^{1}$ yields the lowest workload. This suggests that the expanded dataset contains richer information, allowing global modeling to better capture defect relationships and assign tasks more accurately and efficiently.

\begin{figure}[t]
\graphicspath{{figs/}}
\begin{center}
\includegraphics[width=0.42\textwidth]{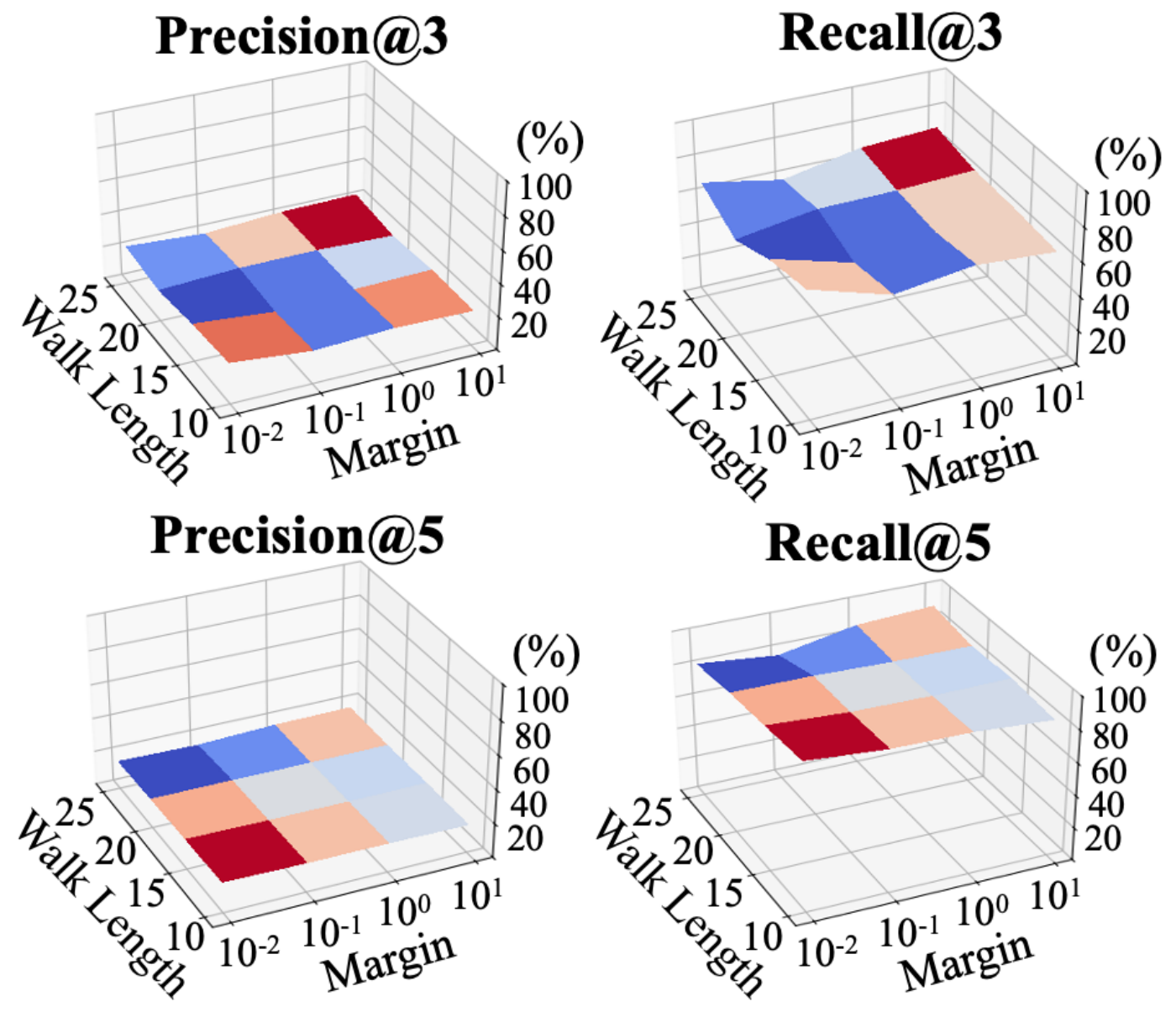}
\end{center}
\caption{The performance of \emph{DeCo} for detect type detection on dataset \textbf{ATE}$_{\text{scar}}$ with $\kappa=1$. Within each metric, higher values are represented in red, while lower values are shown in blue.}
\label{fig:detection}
\end{figure}

\noindent\textbf{Performance on Defect Type Detection.}
We further evaluate \emph{DeCo} for defect type detection using defect-aware representations. In Eq.~\ref{eq:sim}, $\text{prob}(k,p)$ denotes the probability of task $t_k$ belonging to defect type $p$. We select the top-3 and top-5 defect types with the highest $\text{prob}(k,p)$ values as detection results. Figure~\ref{fig:detection} presents the results, analyzing the impact of varying walk length and margin in \emph{DeCo}. Results indicate that performance improves (stronger red) when both walk length and margin are either smallest or largest. Longer walks boost performance when the margin is 10, while a larger margin helps only at maximum walk length. This implies that while longer walks capture more informative structures, they require a larger margin to maintain proper separation in the embedding space. Balancing these two factors can further enhance performance.

\begin{figure}[t]
\graphicspath{{figs/}}
\begin{center}
\includegraphics[width=0.49\textwidth]{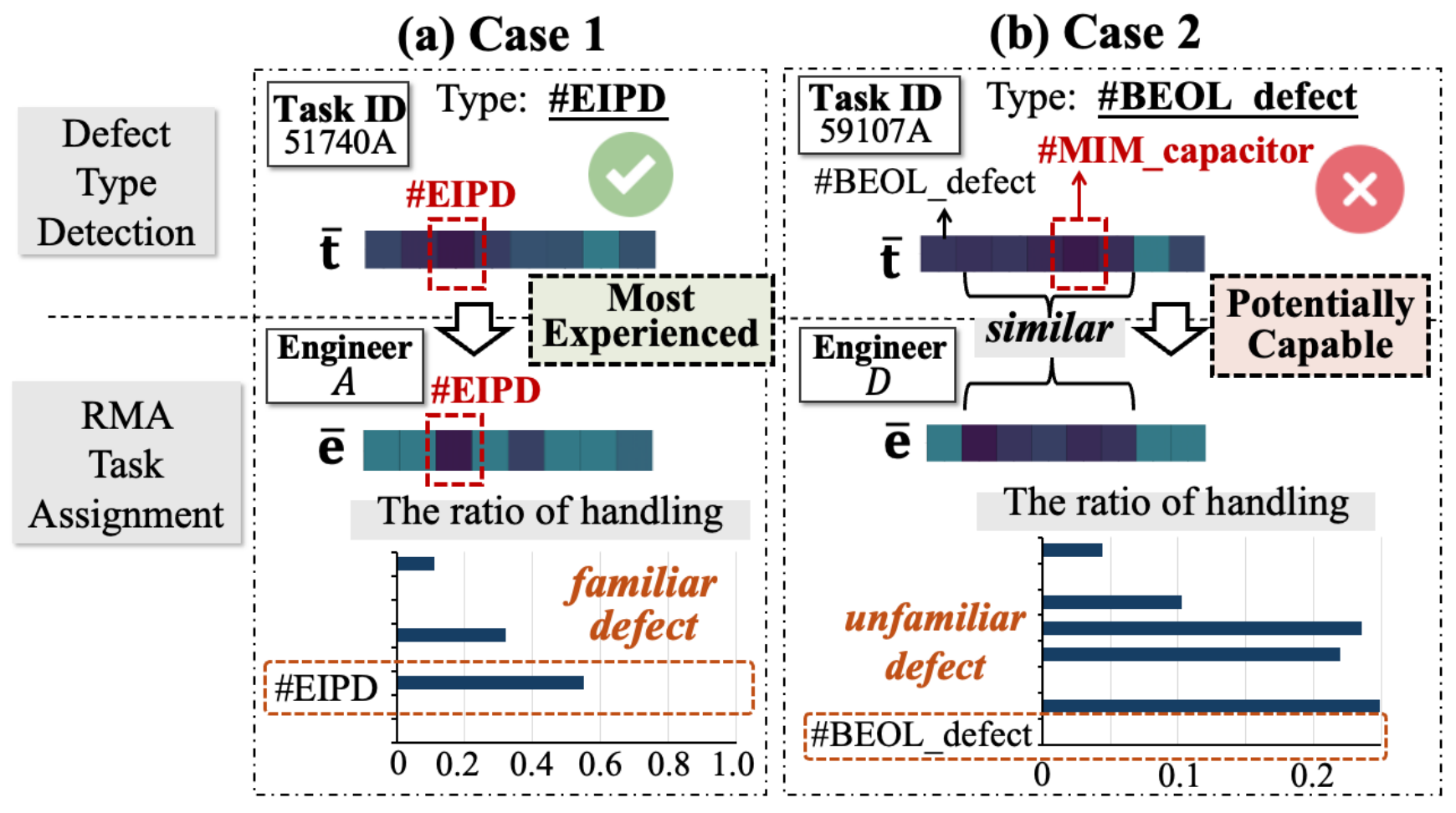}
\end{center}
\caption{Details of two successfully completed cases, analyzed in two stages: defect type detection and RMA task assignment.}
\label{fig:case_study}
\end{figure}

\noindent\textbf{Case study.}
To examine how \emph{DeCo} assigns different tasks, we present two cases where engineers successfully completed tasks before close dates, as shown in Figure~\ref{fig:case_study}. 

\noindent\underline{\textit{Case 1.}} Figure~\ref{fig:case_study}a shows a task of defect type ``\#EIPD''. Based on the learned representation $\bar{\textbf{t}}$ of the task, \emph{DeCo} correctly detects the defect type and assigns the task to Engineer \textit{A}, whose representation $\bar{\textbf{e}}$ indicates the highest capability for handling ``\#EIPD''. Engineer \textit{A} also has the highest historical handling ratio for this defect type. This case demonstrates that defect-aware representations learned in \emph{DeCo} can effectively capture failure characteristics, enabling accurate defect detection and task assignment to the most experienced engineers.

\noindent\underline{\textit{Case 2.}}  Figure~\ref{fig:case_study}b shows a task belonging to type ``\#BEOL\_defect''. As observed in Sec.~\ref{subsec:data_analysis}, ``\#BEOL\_defect'' shares similar failure characteristics with ``\#MIM\_capacitor''. This similarity causes the defect-aware representation $\bar{\textbf{t}}$ to predict a higher probability for ``\#MIM\_capacitor'' than ``\#BEOL\_defect''. However, the contrasting-based assignment in \emph{DeCo} identifies Engineer \textit{D} as a suitable candidate. Even though Engineer \textit{D} has no prior experience with ``\#BEOL\_defect'', the engineer's defect-aware representation $\bar{\textbf{e}}$ is similar to task $\bar{\textbf{t}}$, indicating potential capability. This suggests that Engineer \textit{D}’s experience with other defect types provides transferable skills for handling this task. As a result, \emph{DeCo} assigns the task to Engineer \textit{D}, who successfully completes it despite having no prior experience with this type.

\section{Related Work}

\noindent\textbf{Fault Localization.}
Research on IC testing primarily focuses on optimizing test design in hardware design and manufacturing processes~\cite{9790891,park2017application,afacan2021machine,lyu2018batch}. Machine learning methods have been applied for fault detection and diagnosis during the thin film deposition stage~\cite{9066890}. VAE-based generative models have been used to address data imbalance in thin film deposition~\cite{10035984}. Power supply noise in ICs has also been detected and analyzed using ATE systems \cite{1227027}.

Although existing works perform well in various scenarios, certain defects may exhibit similar failure characteristics, making identification challenging. In such cases, accurate diagnosis often relies on experienced engineers. Therefore, our work focuses on identifying IC failure characteristics and assign them to suitable engineers to improve defect handling. This is an area still unexplored in IC test research.

\noindent\textbf{Task Assignment.}
Recent research focuses on effective assignment while considering constrains~\cite{Mo2013OptimizingPF,hettiachchi2022survey,dickerson2018assigning,andersen2016adaptive,liu2015novel,constantino2017solving}. Some research develops the QASCA system to support quality-aware task assignment by incorporating worker skills and quality standards~\cite{Zheng2015QASCAAQ}. Various evaluation metrics are also used depending on the task type~\cite{Li2014TheWO,lee2014quality,balakrishnan2015task,de2014crowdgrader}. Some studies have assessed the time required for workers to complete tasks~\cite{Mavridis2016UsingHS}, while others have considered fairness among the population under budget constraints~\cite{10.1145/3306618.3314282}.

While these works address worker skills, task difficulty, and efficiency, they often overlook worker workloads, a critical factor in practice. Our RMA task assignment addresses this by considering engineer expertise, task failure characteristics, and workload balance while maximizing success rate and ensuring practical applicability.

\section{Conclusion}

We propose \emph{DeCo}, a framework for optimizing RMA task assignment by constructing a defect-aware graph from ATE logs and employing local and global structure modeling to learn task representations. A contrast-based assignment method refines task-to-engineer allocation. When applied to real-world IC testing data, \emph{DeCo} achieves the highest success rates while keepping engineers' workloads manageable across both scarce and expanded ATE log conditions. We also demonstrate its effectiveness in defect type detection. Two case studies show that \emph{DeCo} can assign potentially capable engineers to previously unfamiliar defects.

\section*{Acknowledgments}

This paper was supported in part by National Science and Technology Council (NSTC), R.O.C., under Contract 113-2221-E-006-203-MY2, 114-2622-8-006-002-TD1 and 113-2634-F-006-001-MBK.

\bibliographystyle{named}
\bibliography{bibliography}

\end{document}